\documentclass[letterpaper, 10 pt, conference]{ieeeconf}
\IEEEoverridecommandlockouts 
\let\labelindent\relax
\usepackage{amsmath, amssymb, booktabs, graphicx, bm, color, algorithm, tabularx, enumitem}
\usepackage[noend]{algpseudocode}
\graphicspath{{images/}}
\PassOptionsToPackage{hyphens}{url}\usepackage{hyperref}

\newcommand{\new}[1]{#1}

\newcolumntype{Y}{>{\centering\arraybackslash}X}

\title{\LARGE \bf Multimodal Material Classification for Robots using \\Spectroscopy and High Resolution Texture Imaging}

\author{Zackory Erickson, Eliot Xing, Bharat Srirangam, Sonia Chernova, and Charles C. Kemp
\thanks{*This work was supported by NSF award IIS-1514258 and AWS Cloud Credits for Research. Dr. Kemp owns equity in and works for Hello Robot, a company commercializing robotic assistance technologies.}%
\thanks{Zackory Erickson, Eliot Xing, Bharat Srirangam, and Charles C. Kemp are with the Healthcare Robotics Lab, Georgia Institute of Technology, Atlanta, GA., USA.}%
\thanks{Sonia Chernova is with the Robot Autonomy and Interactive Learning Lab, Georgia Institute of Technology, Atlanta, GA., USA.}%
\thanks{Zackory Erickson is the corresponding author {\tt\footnotesize zackory@gatech.edu}.}%
}

\begin{document}

\maketitle
\thispagestyle{empty}
\pagestyle{empty}

\begin{abstract}
Material recognition can help inform robots about how to properly interact with and manipulate real-world objects. In this paper, we present a multimodal sensing technique, leveraging near-infrared spectroscopy and close-range high resolution texture imaging, that enables robots to estimate the materials of household objects. We release a dataset of high resolution texture images and spectral measurements collected from a mobile manipulator that interacted with 144 household objects. We then present a neural network architecture that learns a compact multimodal representation of spectral measurements and texture images. When generalizing material classification to new objects, we show that this multimodal representation enables a robot to recognize materials with greater performance as compared to prior state-of-the-art approaches. Finally, we present how a robot can combine this high resolution local sensing with images from the robot's head-mounted camera to achieve accurate material classification over a scene of objects on a table.
\end{abstract}

\section{Introduction}
\label{sec:intro}

When interacting with everyday objects, people frequently use material properties to inform their interactions~\cite{buckingham2009cueslifting}.
We make sure not to place metal in the microwave, we take caution when carrying glass or ceramic objects, we look for styrofoam or paper cups to hold hot liquids, and we sort some paper, plastic, and metal objects into recycling bins.
Robots can benefit from these same skills when operating in human environments.

In this work, we demonstrate how robots can use a non-contact multimodal sensing technique, based on spectroscopy and close-range texture imaging, to accurately estimate the materials of household objects prior to manipulation.
This sensing approach collects near-infrared spectral measurements from a handheld micro spectrometer with a narrow field-of view camera for high resolution texture imaging. Both sensors are small and can be held by or directly integrated into a robot's end effector.
Non-contact sensing can enable a robot to determine properties and use cases of objects without the intricacies of contact physics that can affect the performance of haptic touch-based sensing.

\begin{figure}
\vspace{0.2cm}
\centering
\includegraphics[width=0.47\textwidth, trim={0cm 0cm 0cm 5cm}, clip]{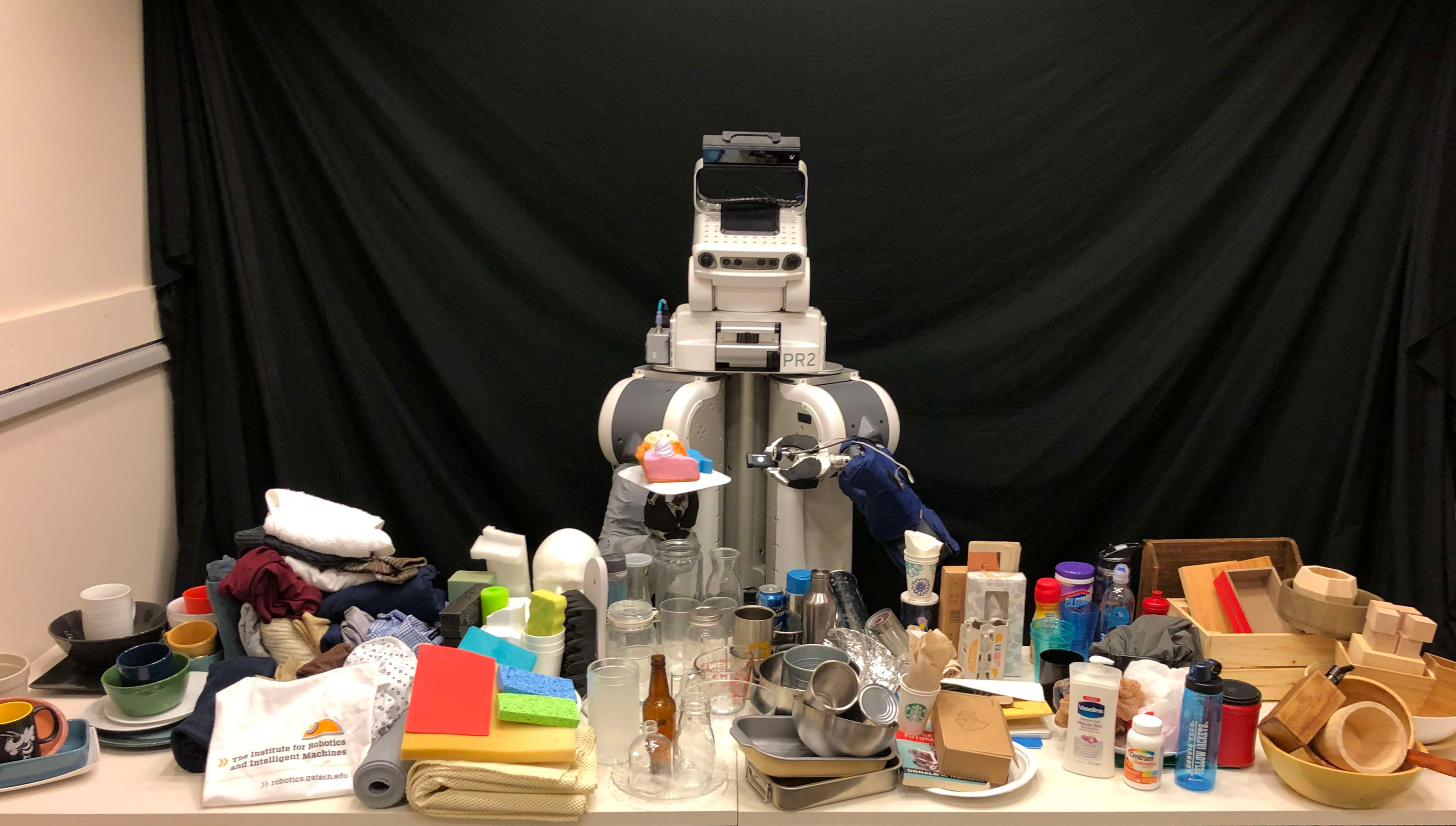}
\caption{\label{fig:intro}The PR2 used a spectrometer and near-field camera to estimate the material of 144 everyday objects.}
\vspace{-0.4cm}
\end{figure}

To evaluate this multimodal sensing technique, we have assembled and released a dataset of 14,400 high resolution texture images and corresponding spectral measurements. 
We collected this data with a PR2 mobile manipulator that interacted with 144 household objects, shown in Fig.~\ref{fig:intro}, which spanned eight material categories: ceramic, fabric, foam, glass, metal, paper, plastic, and wood.

Using this dataset, we trained a neural network that learns a shared representation of spectral and visual sensory data.
By learning a compact multimodal representation, our model achieves state-of-the-art material recognition performance of 80.0\% when generalizing material classification to a new set of heldout objects across eight materials (12.5\% baseline with a random classifier). 
We further investigate the role of texture image preprocessing by comparing several ImageNet-pretrained CNN models for generating lower-dimensional visual representations.
Finally, using this spectral and visual sensing approach, we demonstrate that a robot can reliably classify a scene of objects on a table without direct contact. In this work, we make the following contributions:
\begin{itemize}[noitemsep,topsep=0pt]
\item We introduce a near-infrared spectroscopy and high resolution texture imaging approach that surpasses prior state-of-the-art performance~\cite{erickson2019spec} for material classification.
\item We release SpectroVision, a dataset of 14,400 high resolution texture images and spectral measurements collected from a PR2 mobile manipulator that interacted with 144 household objects from eight material categories.
\item We demonstrate that our multimodal approach surpasses the performance of models trained on each independent modality.
\item We show that a robot equipped only with our handheld sensors and an RGB-D camera can successfully use our approach to perform material classification on multiple objects casually arranged on a table.
\end{itemize}

\section{Related Work and Background}
\label{sec:related_work}

\subsection{Material Recognition}

Material recognition using haptic sensors, which require direct physical contact with objects, has been widely explored. Modalities such as force~\cite{bhattacharjee2012tactile, decherchi2011tactile}, temperature~\cite{kerr2013thermal, bhattacharjee2015heat, cho2018thermal}, capacitance~\cite{alagi2018capacitive}, vibration~\cite{sinapov2011vibrotactile, fang2019acoustic}, and radar~\cite{yeo2016radarcat}, have been used in haptic perception for material recognition. The BioTac fingertip, capable of sensing force, temperature, and vibration, has been studied for multimodal haptic perception~\cite{chu2015hapticadjectives, fishel2012biotac, xu2013tactile, kerr2018biotac}. Chin et al. introduced a compliant haptic sensor for robots to distinguish between plastic, metal, and paper during recycling~\cite{chin2019automated}. Several works also use multimodal perception by combining data from multiple modalities for material recognition and outperforming single modality approaches~\cite{chathuranga2013investigation, sinapov2014grounding, erickson2017semi, bhattacharjee2018multimodal, zhang2019multimodalcutting}.
Similarly, we find that non-contact material recognition approaches also benefit from multiple sensing modalities, and we demonstrate that visual sensing couples well with spectroscopy.

Several studies have evaluated visual features for material recognition~\cite{liu2010exploring, hu2011toward, dimitrov2014vision, bell2015material, schwartz2019recognizing, su2016material, wang20164d}.
Extensive literature also exists for leveraging visual or depth imaging for vision-based tactile sensors, including the GelSight~\cite{yuan2017shape, li2013sensing}, FingerVision~\cite{yamaguchi2017implementing}, and TacTip~\cite{ward2018tactip, cramphorn2018voronoi}, to perform manipulation tasks~\cite{chen2018tactile, yamaguchi2016combining}, texture recognition~\cite{luo2018vitac, lee2019touching}, and estimation of material properties~\cite{kampouris2016fine, yuan2017gelsight}.
Both \cite{yuan2017shape} and \cite{gao2016deep} have used visual and haptic features to estimate object properties, such as hardness or haptic adjectives. Overall, we find that multimodal approaches overcome weaknesses in the ability of any individual modality to classify materials. 

\subsection{Spectroscopy}
Spectroscopy~\cite{pasquini2018specreview} has found a number of practical applications such as for pharmaceutical manufacturing~\cite{roggo2007pharma}, food analysis~\cite{bellon1994food}, and recycled material separation~\cite{masoumi2012plastic}. Recently, a number of handheld spectrometers have been developed for performing spectral analysis outside of lab and manufacturing settings~\cite{crocombe2018portable, rateni2017smartphonebased}. These portable micro spectrometers have been demonstrated for pharmaceutical quality control~\cite{yan2018pharma} and food analysis~\cite{das2016fruit, lee2017nir, kartakoullis2018meat}. 

Prior research has shown how a robot can use near-infrared spectroscopy with a commercial handheld SCiO spectrometer to recognize materials of household objects~\cite{erickson2019spec}.
Near-infrared spectroscopy has since been used by robots to recognize the materials of household objects for informing semantic grasp predictions and for tool construction~\cite{liu2019cage, nair2019autonomous, shrivatsav2019tool}.
In this paper, we demonstrate that robots can more accurately recognize common household materials by leveraging both spectroscopy and close-range texture imaging.

\subsection{Texture Representation}

Several techniques have been introduced for extracting or learning texture representations from visual images, including convolutional neural network (CNN) based texture analysis~\cite{liu2019bow} and handcrafted descriptors~\cite{paolo2017texturedescriptors}.
Recent work in texture analysis has primarily investigated CNN-based texture representations~\cite{liu2019bow}. This is due in part to a collection of works in texture and material classification tasks that have shown learned CNN feature descriptors frequently outperform alternative, handcrafted approaches~\cite{bell2015material, schwartz2019recognizing, liu2019bow, paolo2017texturedescriptors, kalliatakis2017evaluating}.

Research in texture synthesis~\cite{gatys2015texture, lin2016visualizing} has also provided insight into the ways in which CNNs capture and encode textures.
Vision-based tactile sensing techniques for texture classification have frequently used texture features from pretrained ImageNet models~\cite{yuan2018active}.
The use of these models for extracting textural features is further supported by findings of Geirhos et al.~\cite{geirhos2018imagenettexture} that ImageNet-trained CNNs are more biased towards recognizing and representing localized textures rather than global shape structure, similar to results by~\cite{gatys2015texture, long2018texturestat, ballester2016cnnsketches}.
Building on these prior findings, we leverage pretrained ImageNet CNNs to extract robust visual texture features for material classification.

\section{SpectroVision Dataset}
\label{sec:dataset}
\begin{figure}
\centering
\includegraphics[width=0.23\textwidth, trim={3cm 3cm 0cm 0cm}, clip]{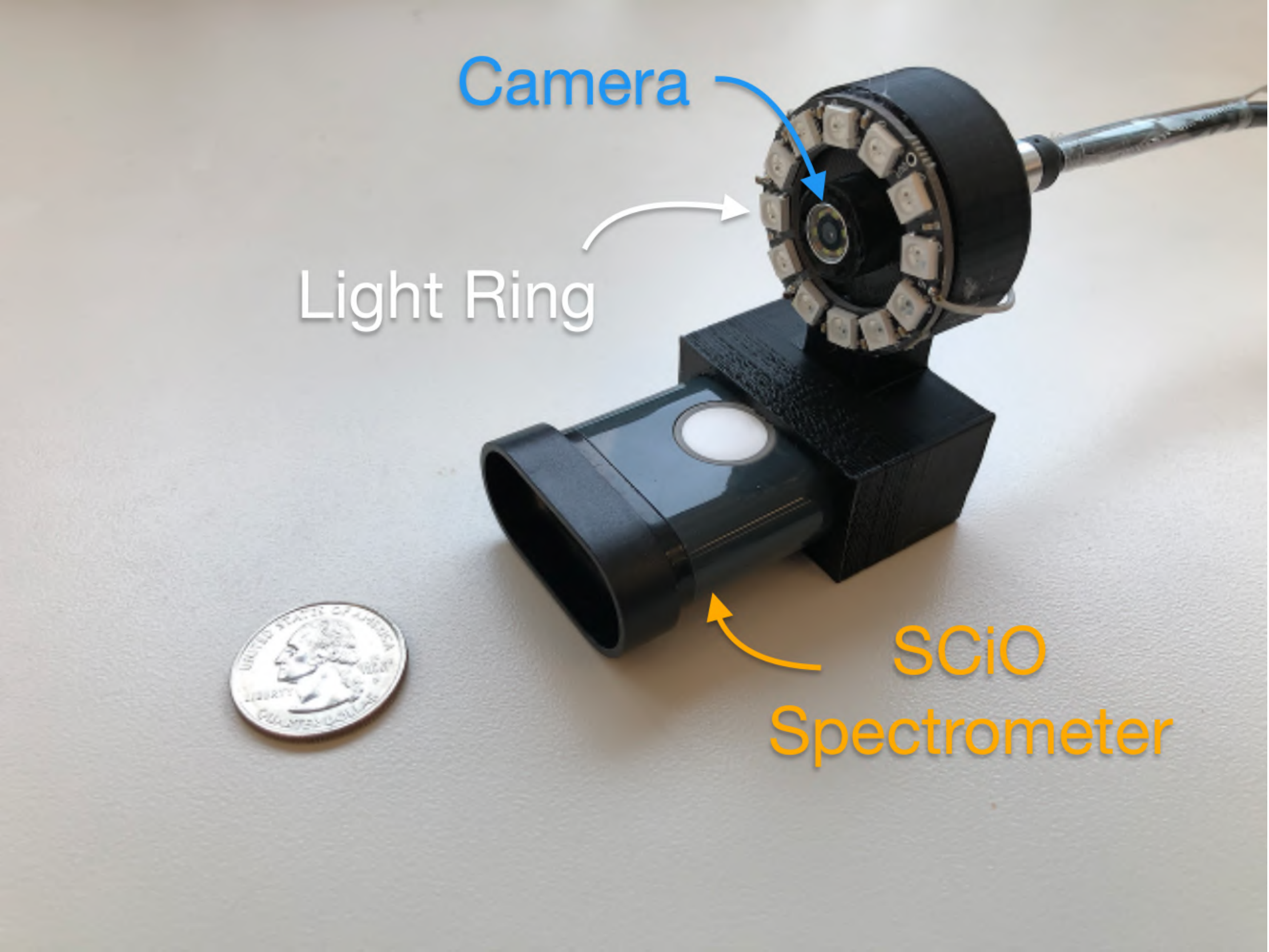}
\includegraphics[width=0.23\textwidth, trim={0cm 0cm 0cm 1.9cm}, clip]{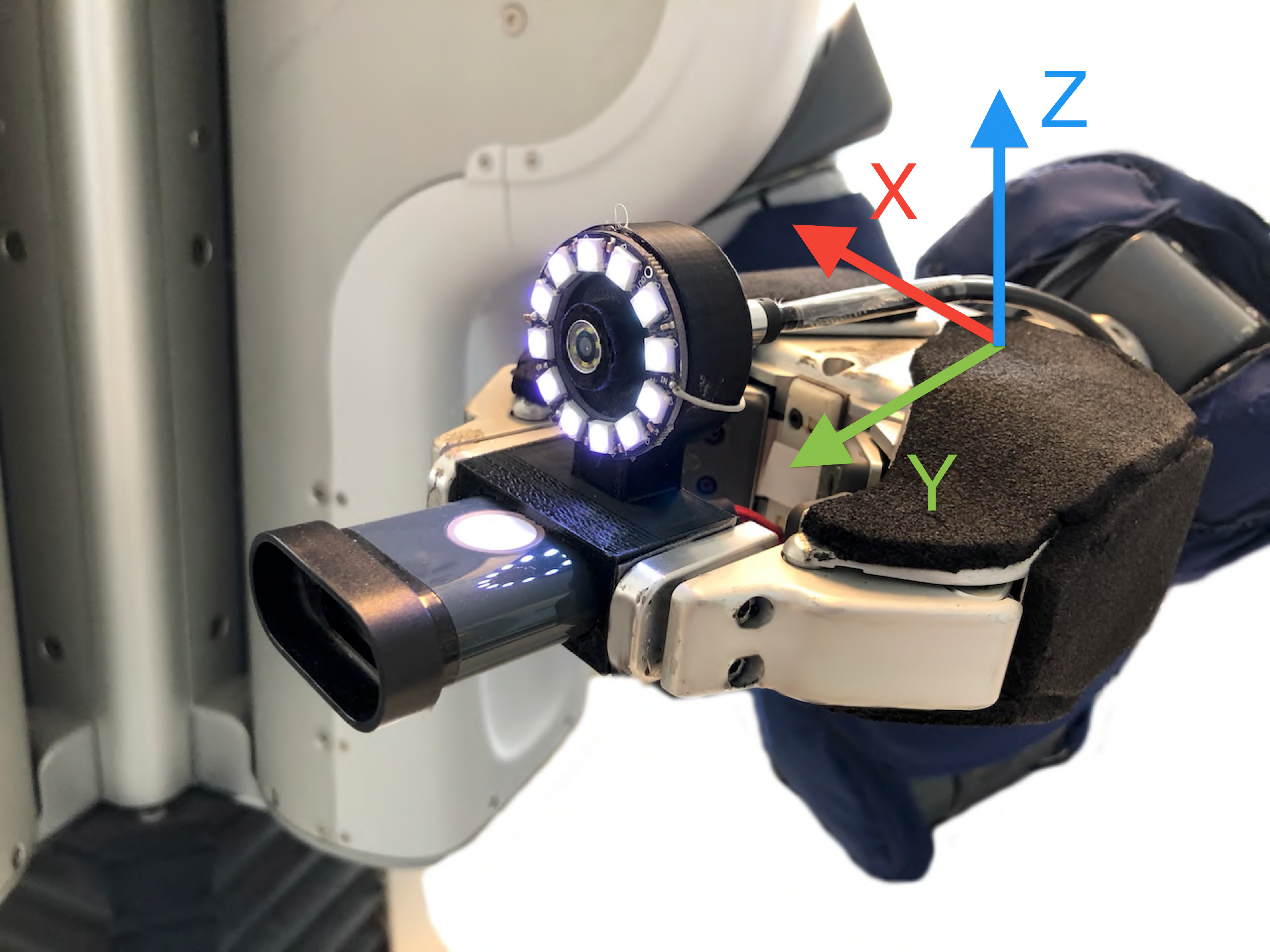}
\vspace{-0.1cm}
\caption{\label{fig:sensors}Figures of the SCiO and camera/light ring sensor setup. (Left) On a table, with a quarter shown for sizing. (Right) Held by the PR2.}
\vspace{-0.3cm}
\end{figure}

\subsection{Sensors}

Our sensing approach consists of a micro handheld spectrometer for near-infrared spectral measurements and a narrow field-of view camera for high resolution texture imaging.
Compared to haptic sensing, spectroscopy and imaging have advantageous properties for material recognition, including fast response times and no physical contact requirements.

Fig.~\ref{fig:sensors} shows the SCiO spectrometer and camera, by themselves and when held in a PR2 robot's end effector. The SCiO is a near-infrared spectrometer that measures light spectra in the wavelength range of $\lambda=$~740~nm to $\lambda=$~1,070~nm. 
The 35 gram spectrometer is Bluetooth enabled and has a black pigmented cover around the sensor aperture which ensures there is an $\sim$1~cm minimum air gap between an object and the sensor aperture.

We capture texture images with a 2~megapixel endoscope camera. The 8.4~mm diameter camera has an optimal viewing distance of 6~cm to 10~cm and is capable of capturing images at 1600 $\times$ 1200 resolution. We placed a 12 LED light ring around the camera (see Fig.~\ref{fig:sensors}) to ensure consistent illumination of each object that the robot interacts with. 

We attached the spectrometer and camera together with a grasping mount for the PR2's end effector. Note that there is an $\sim$3.5cm offset between the apertures of the two sensors.

\begin{figure}
\centering
\includegraphics[width=0.23\textwidth, trim={5cm 4cm 5cm 1cm}, clip]{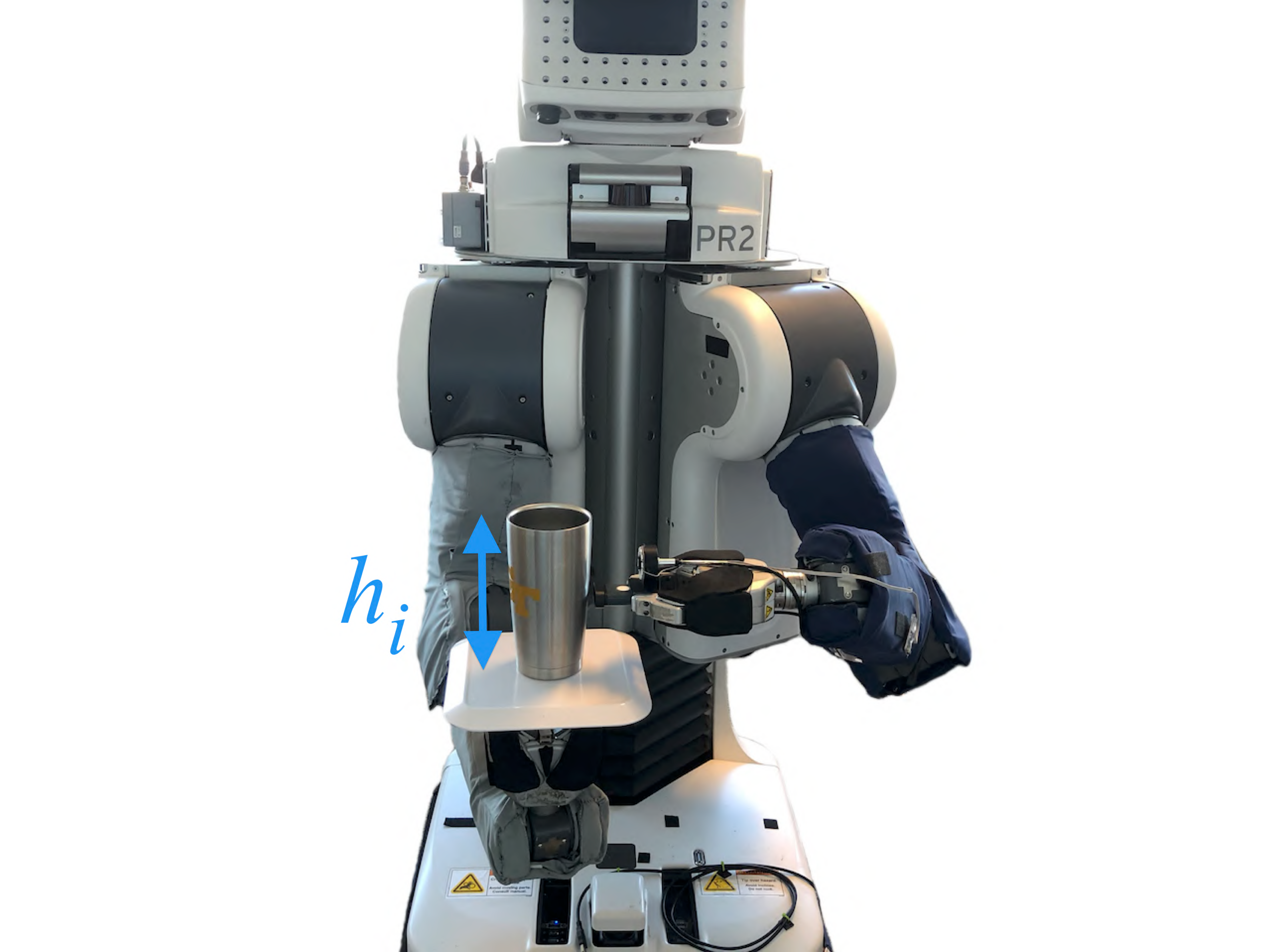}
\includegraphics[width=0.23\textwidth, trim={6.2cm 5cm 6.2cm 2cm}, clip]{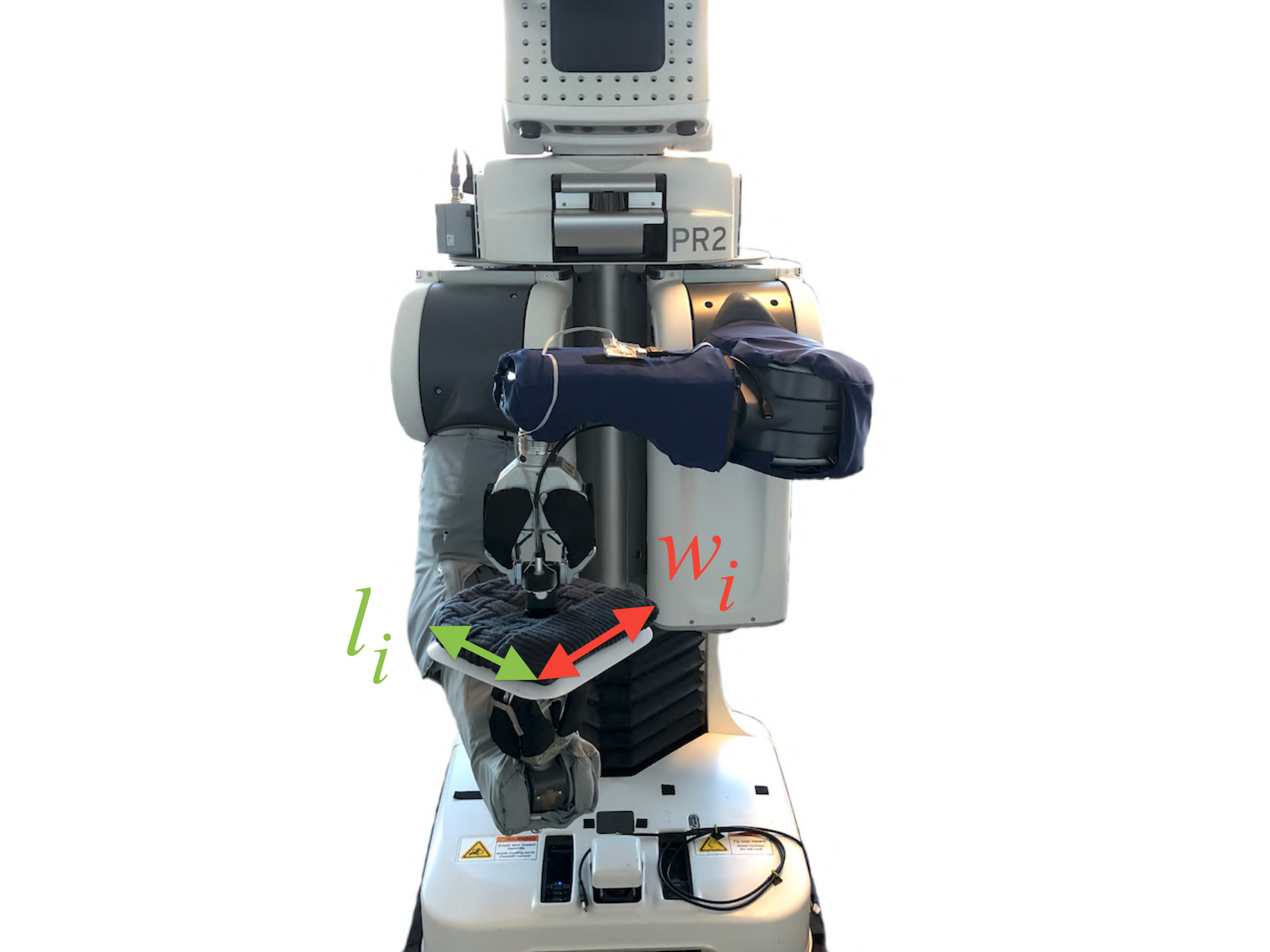}
\vspace{-0.1cm}
\caption{\label{fig:datacollection}Demonstration of data collection with the PR2. (Left) Interaction with a vertical object. (Right) Horizontal object interaction.}
\vspace{-0.4cm}
\end{figure}

\begin{figure*}
\centering
\includegraphics[width=0.1175\textwidth, trim={0cm 0cm 0cm 0cm}, clip]{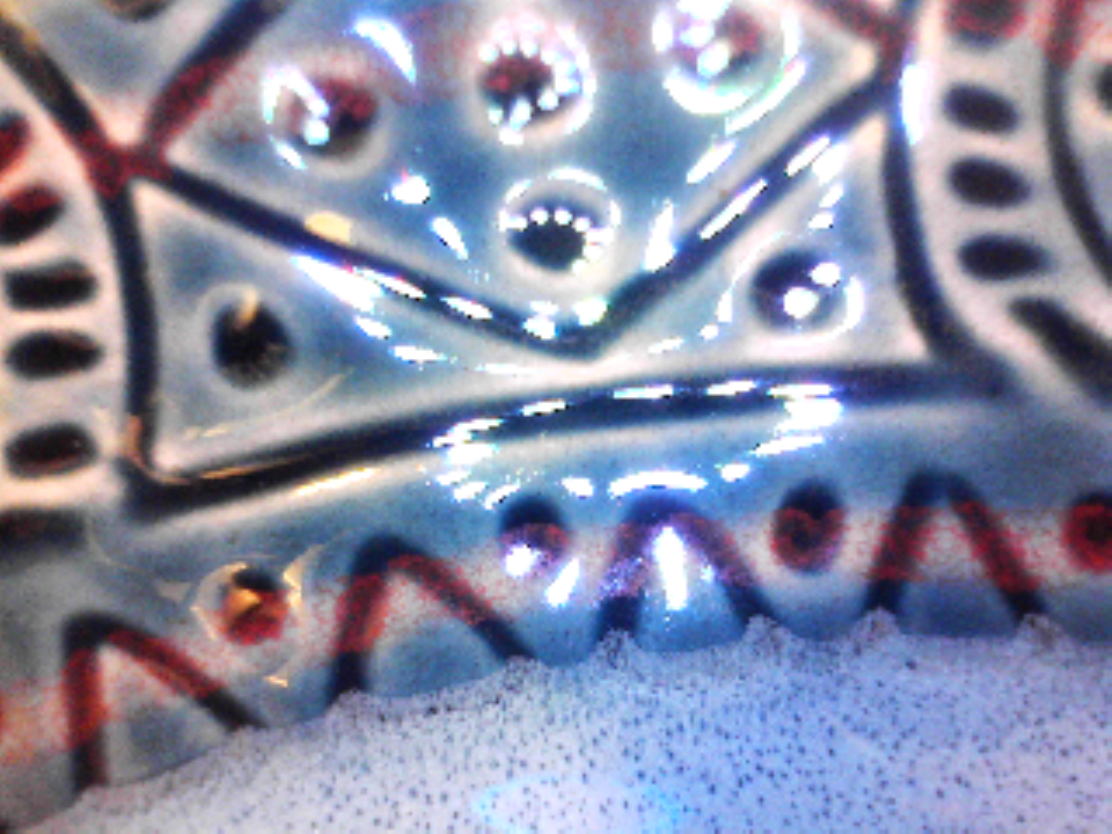}
\includegraphics[width=0.1175\textwidth, trim={0cm 0cm 0cm 0cm}, clip]{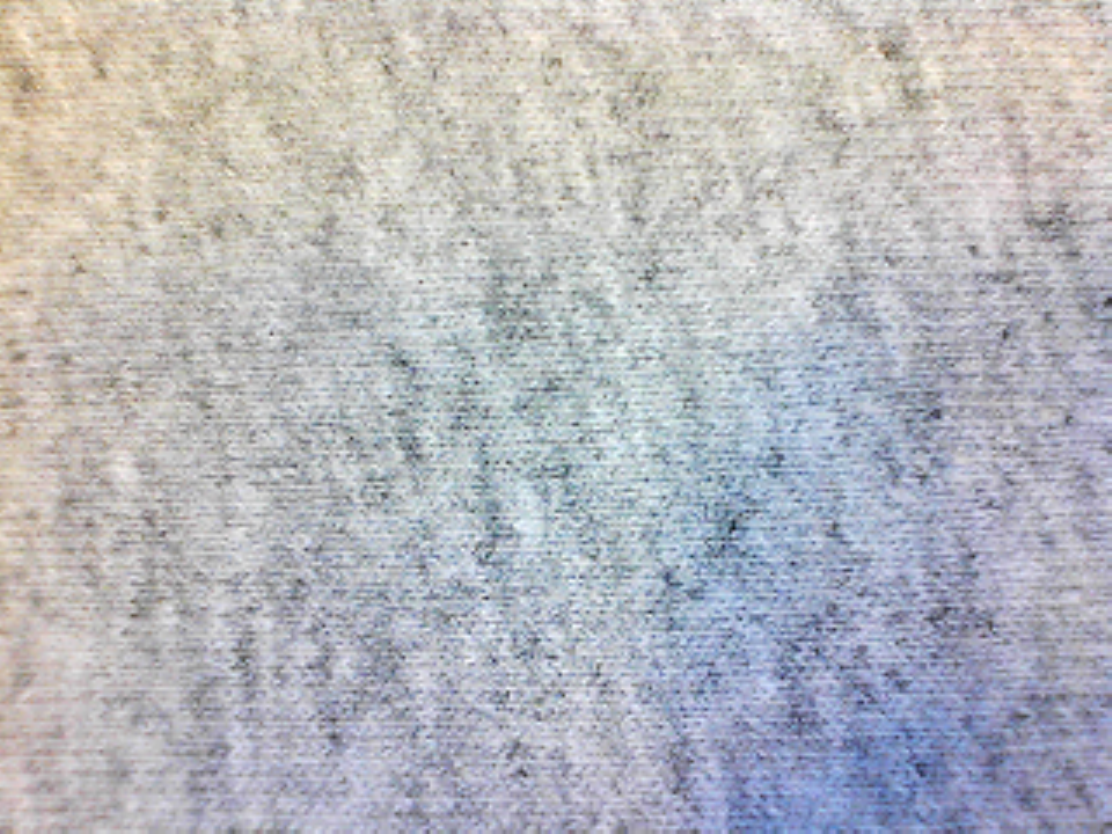}
\includegraphics[width=0.1175\textwidth, trim={0cm 0cm 0cm 0cm}, clip]{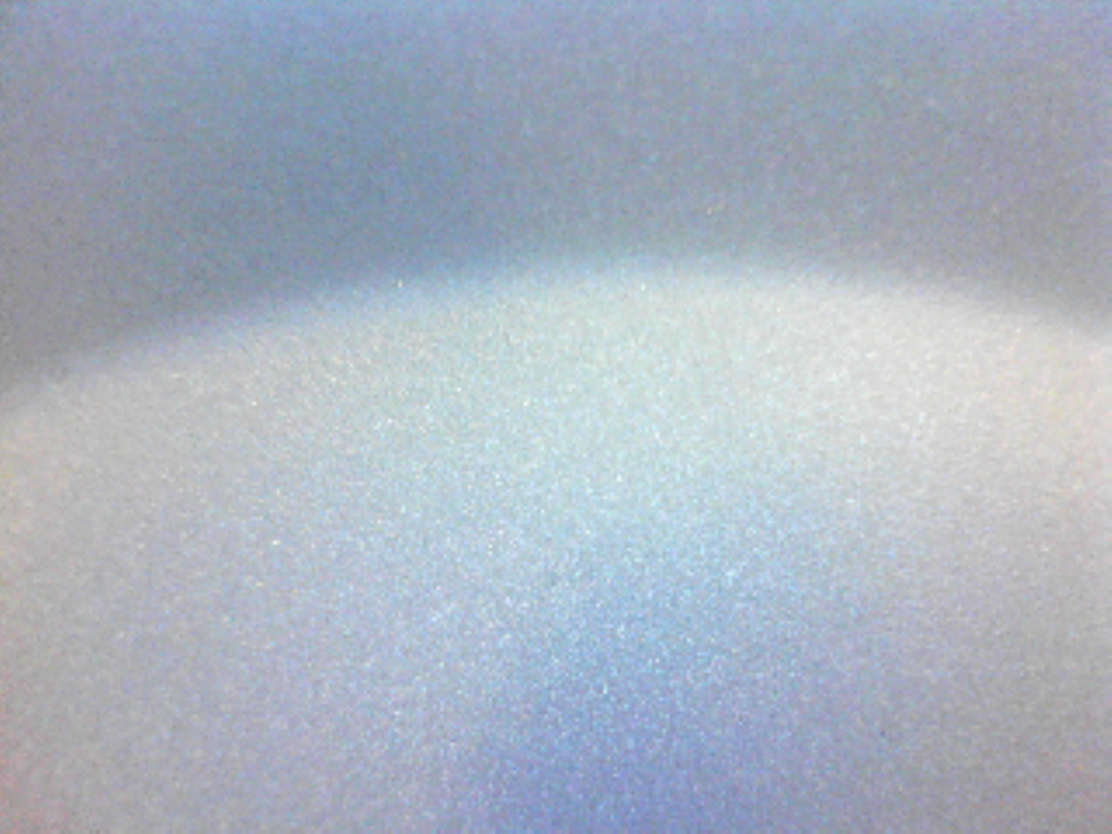}
\includegraphics[width=0.1175\textwidth, trim={0cm 0cm 0cm 0cm}, clip]{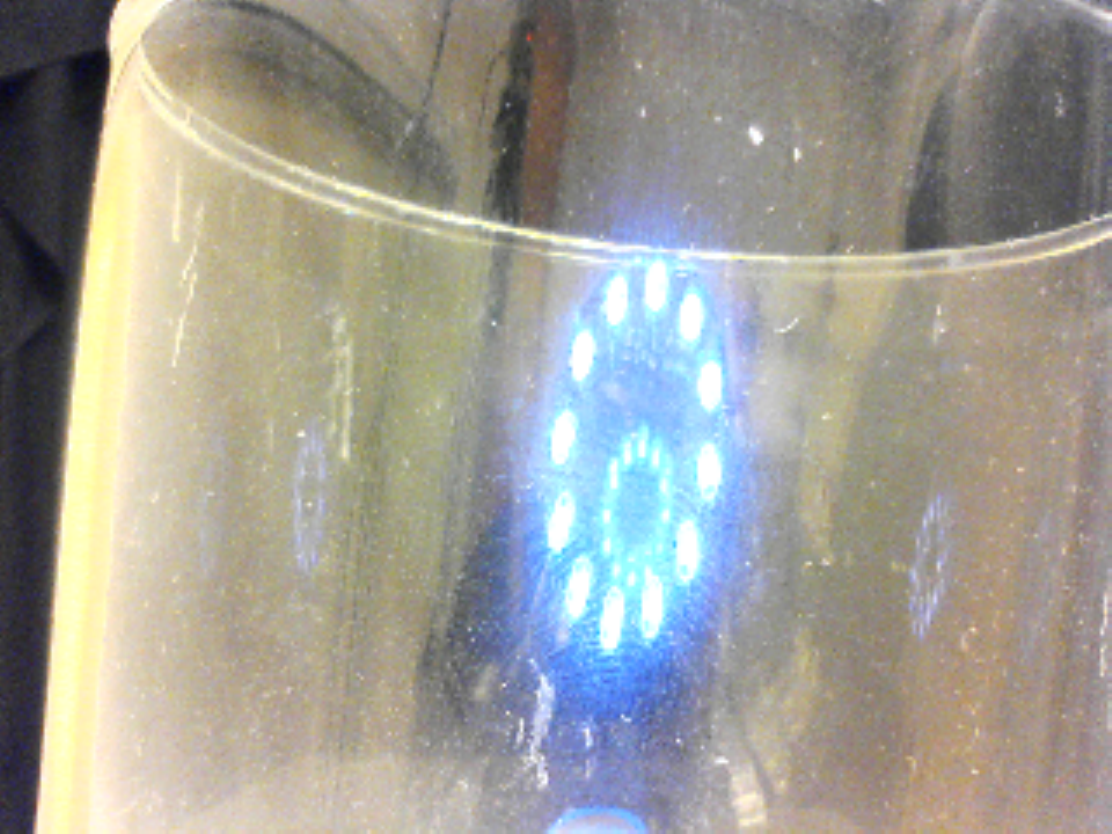}
\includegraphics[width=0.1175\textwidth, trim={0cm 0cm 0cm 0cm}, clip]{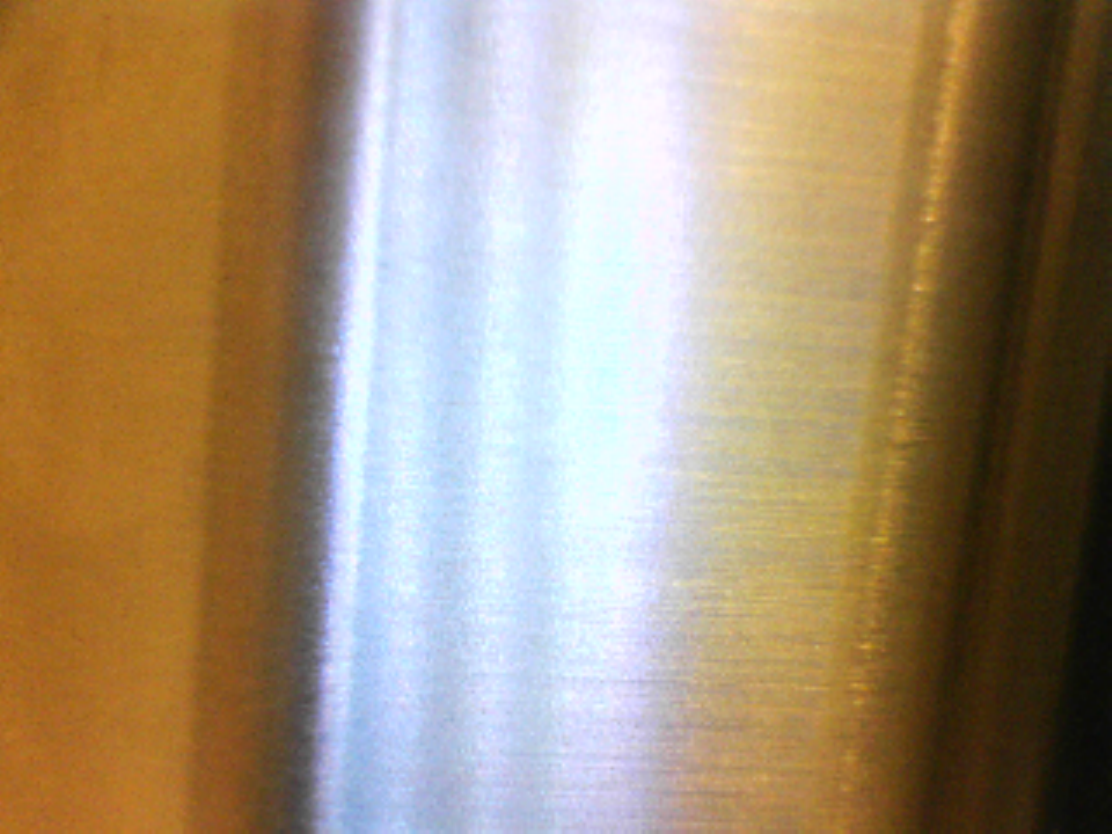}
\includegraphics[width=0.1175\textwidth, trim={0cm 0cm 0cm 0cm}, clip]{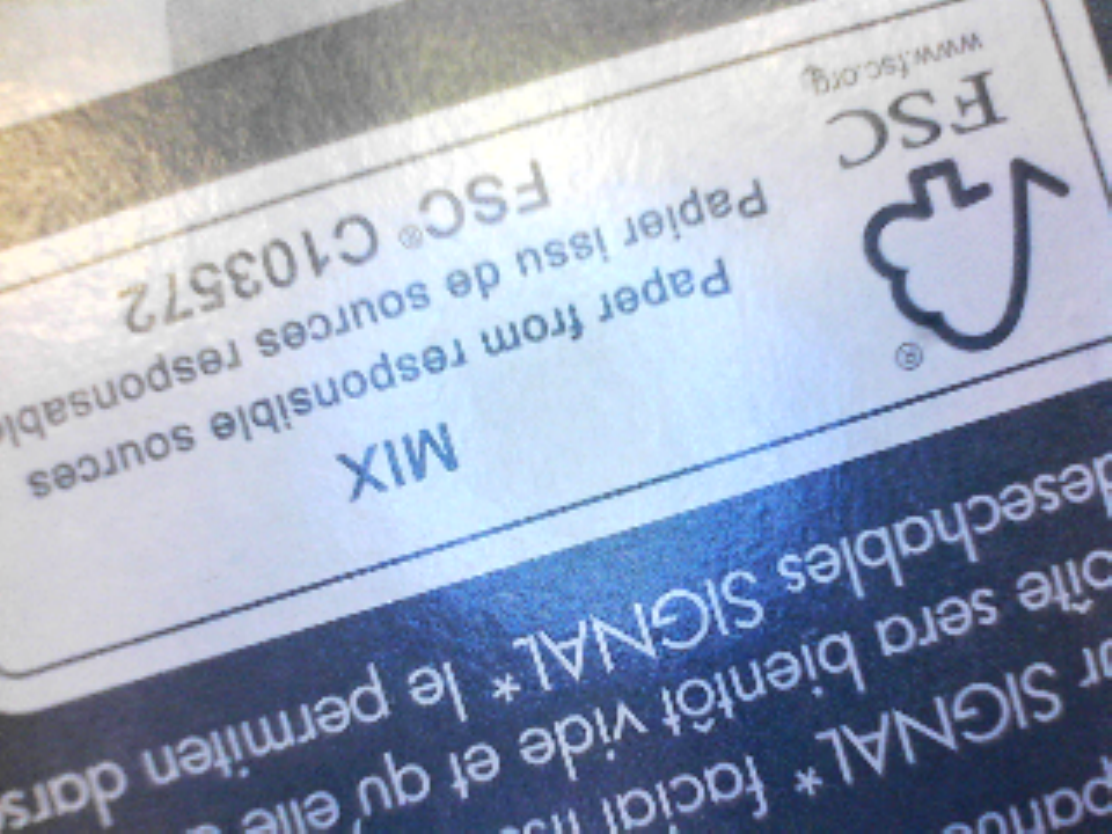}
\includegraphics[width=0.1175\textwidth, trim={0cm 0cm 0cm 0cm}, clip]{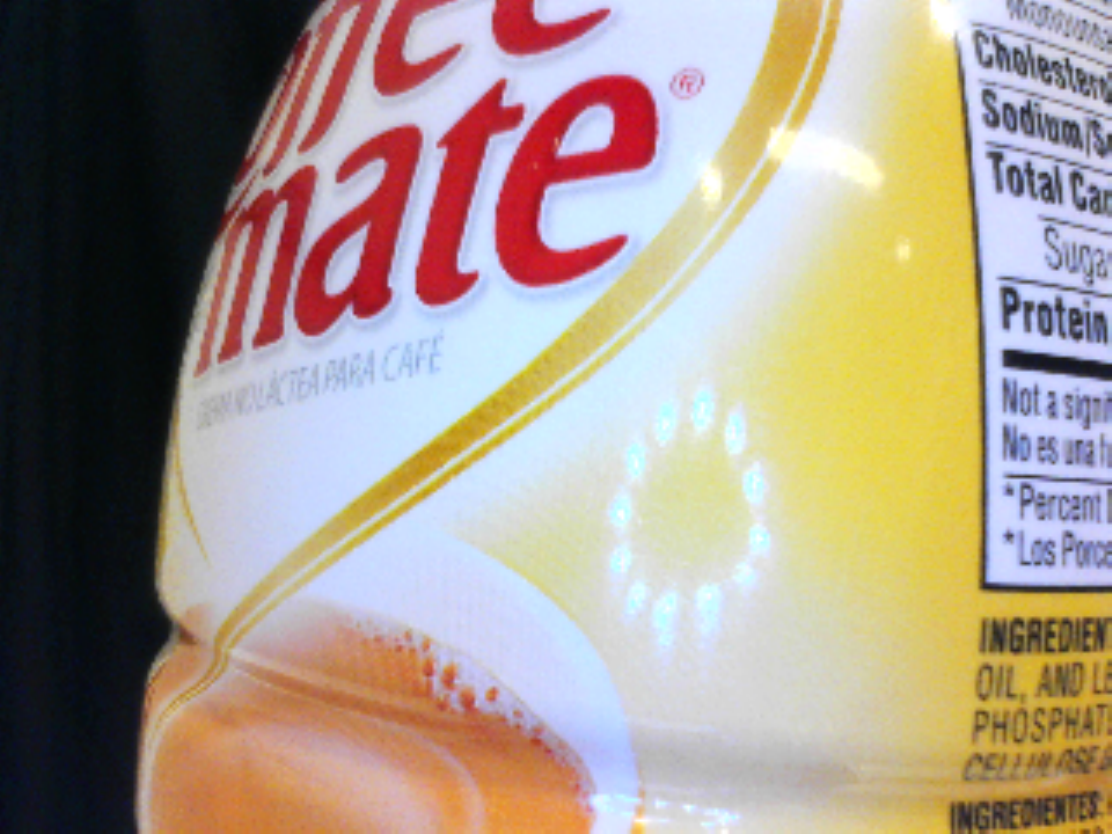}
\includegraphics[width=0.1175\textwidth, trim={0cm 0cm 0cm 0cm}, clip]{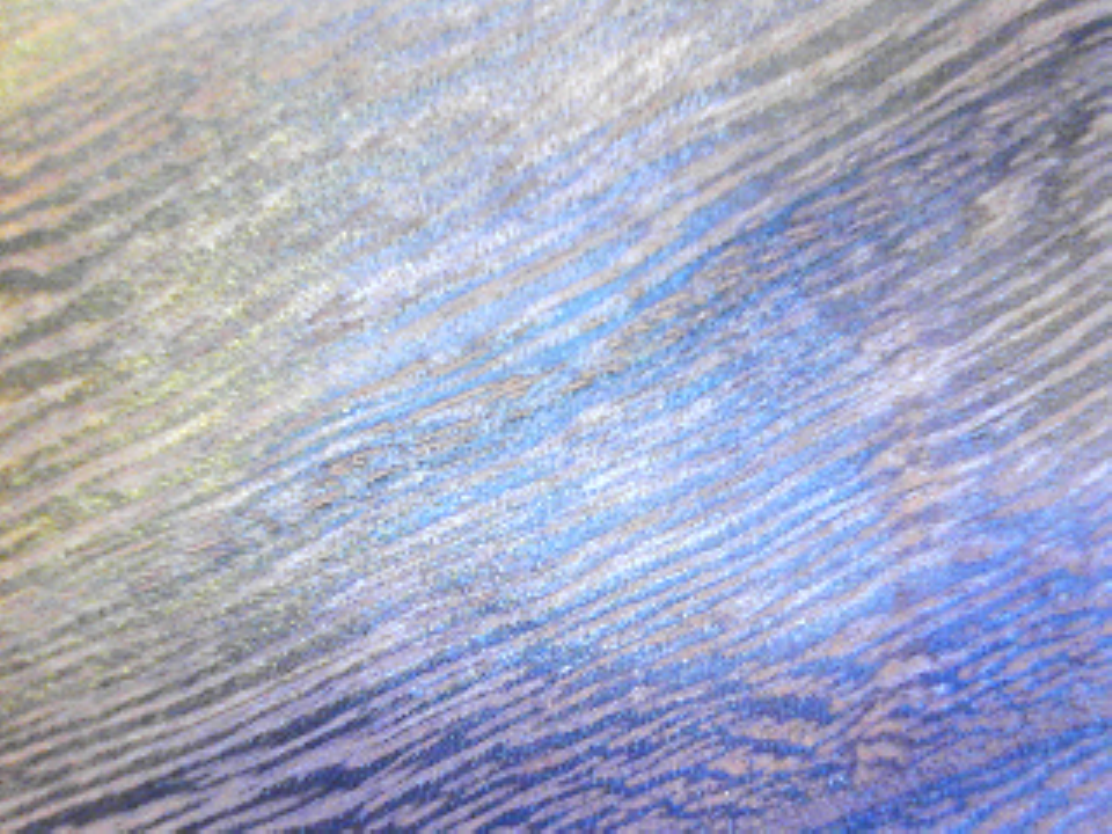}\\
\begin{tabularx}{\textwidth} { Y Y Y Y Y Y Y Y }
    Ceramic & Fabric & Foam & Glass & Metal & Paper & Plastic & Wood
\end{tabularx}
\vspace{-0.6cm}
\caption{\label{fig:imagewall}Examples of texture images from each material category.}
\vspace{-0.4cm}
\end{figure*}

\begin{figure}
\centering
\includegraphics[width=0.47\textwidth, trim={0cm 0cm 0cm 0cm}, clip]{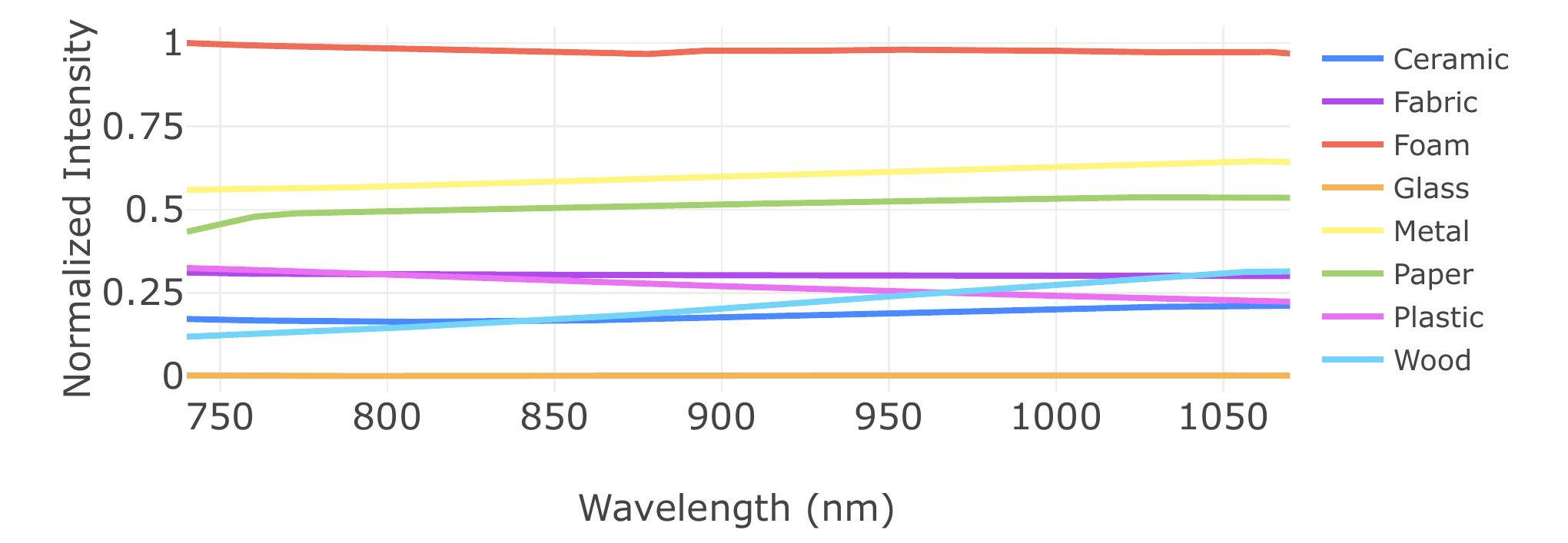}
\vspace{-0.2cm}
\caption{\label{fig:scio_measurements}Example raw spectral samples for each of the objects in Fig.~\ref{fig:imagewall}.}
\vspace{-0.4cm}
\end{figure}

\subsection{Dataset and Data Collection}

We have collected and released SpectroVision, a dataset\footnote{SpectroVision\:dataset:\:\url{ https://github.com/Healthcare-Robotics/spectrovision/releases}} of 14,400 texture images and near-infrared spectral samples. This data was captured from a PR2 robot that interacted with 144 household objects from 8 material categories, as shown in Fig.~\ref{fig:intro}. These materials include ceramic, fabric, foam, glass, metal, paper, plastic, and wood, with 18 unique objects per material.

The robot performed 100 interactions with each object, sampling at random positions and orientations along an object's outer surface. To do this, the robot used its right end effector to hold a flat platter on which we rigidly mounted objects to, as demonstrated in Fig.~\ref{fig:datacollection}. For measurements collected with vertically standing upright objects, the robot would rotate the platter, then randomly sample a roll orientation for the left end effector \new{$\theta_y \in [-\frac{\pi}{9}, \frac{\pi}{9}]$ (see Fig.~\ref{fig:sensors})} and a vertical height to interact with the object at in $[0, h_i]$, where $h_i$ represents the height of object $i$ \new{(see Fig.~\ref{fig:datacollection})}. For objects that lie flat on the platter, the robot would randomly sample an end effector roll orientation $\theta_y \in [-\frac{\pi}{6}, \frac{\pi}{6}]$ and a point of contact in $[0, l_i]$, $[0, w_i]$ along the top surface of the object, with length $l_i$ and width $w_i$. Due to the random roll orientation of the robot's end effector and the $\sim$3.5~cm height offset between the spectrometer and camera (seen in Fig.~\ref{fig:sensors}), spectral and texture images captured at the same time are not co-located and hence pairings between these measurements are not strict. In early evaluations, we found that randomizing pairings between spectral and image samples from the same object did not have considerable impact on classification performance.
Video sequences of the data collection process can be found in the supplementary video.
\new{We note that future research could extend the results in this work by generalizing to other common object sets~\cite{calli2017yale}, or evaluating multilabel classification with objects of non-homogeneous materials.}

In comparison to some haptic sensing approaches that can take upwards of 15-20 seconds per measurement~\cite{sinapov2011vibrotactile, kerr2018biotac}, spectroscopy and imaging offer consistently fast sensing times. Capturing an image takes $\sim$1.5 milliseconds, whereas the SCiO has a 1-2 second sensing time, which consists of $\sim$1 second of light exposure, reflectance data processing, and Bluetooth communication. 
Data processing consists of normalizing the raw spectrum reading from the SCiO's optical head by the raw spectrum of a calibration apparatus (a high reflectance mirror material).

Fig.~\ref{fig:imagewall} depicts sample images from each material category, captured by the camera during the interactions. Fig.~\ref{fig:scio_measurements} shows the spectral measurements, which were captured alongside the images in Fig.~\ref{fig:imagewall}. A raw spectral measurement consists of a 331-dimensional vector with a 1~nm wavelength step between the range of $\lambda=$~740~nm to $\lambda=$~1,070~nm. Prior works have shown that the difference quotient (numerical first order derivative) of spectral measurements can improve learning performance~\cite{erickson2019spec, strother2009nir}. Given this finding, we concatenate the difference quotient to each raw spectral measurement, resulting in a 662-dimensional spectral vector.

\subsection{Multimodal Learning Architecture}
\label{ssec:models}

\begin{figure}
\centering
\begin{tabular}{p{0.1cm} l}
(A) & \raisebox{-.5\height}{\includegraphics[width=0.45\textwidth, trim={0cm 0cm 0cm 0cm}, clip]{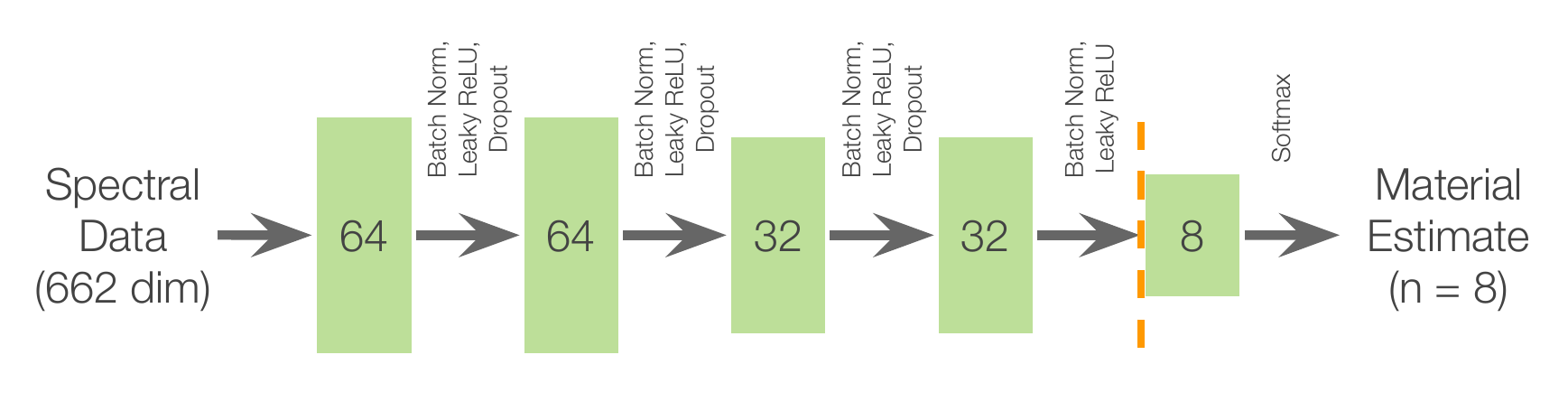}} \\
(B) & \raisebox{-.5\height}{\includegraphics[width=0.45\textwidth, trim={0cm 0cm 0cm 0cm}, clip]{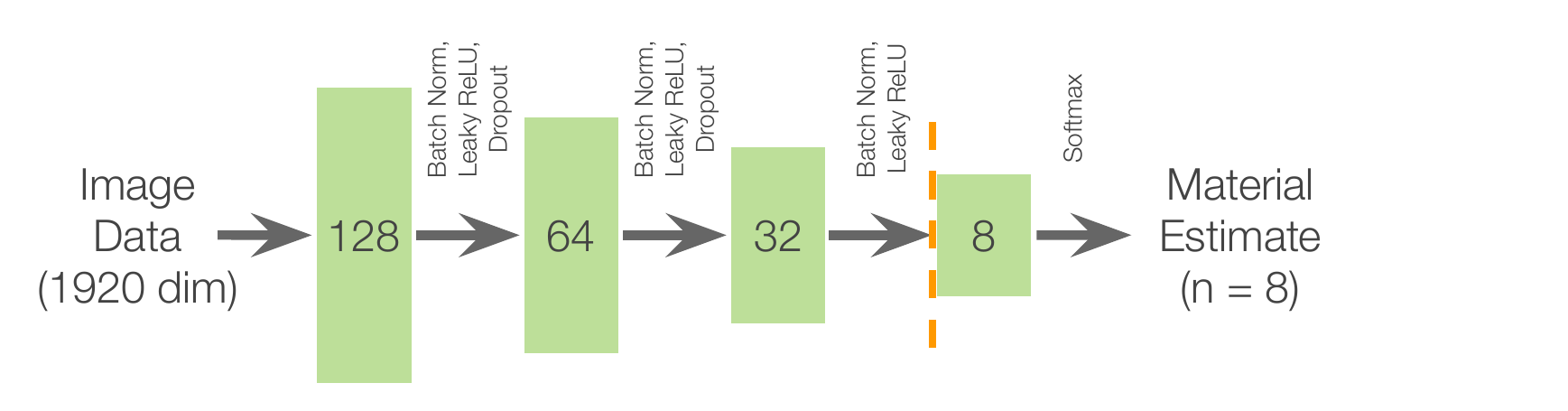}} \\
(C) & \raisebox{-.5\height}{\includegraphics[width=0.45\textwidth, trim={0cm 0cm 0cm 0cm}, clip]{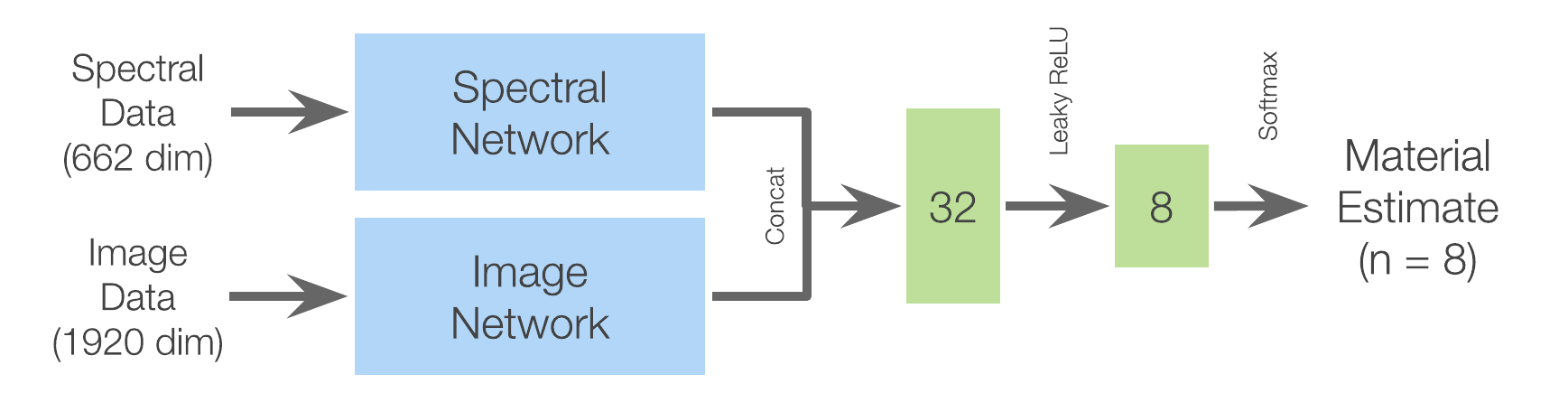}} \\
\end{tabular}
\vspace{-0.2cm}
\caption{\label{fig:networks}The spectral, image, and multimodal network architectures. For the multimodal network (C), the spectral and image networks are first pretrained and then trimmed at the dashed orange line shown above.}
\vspace{-0.5cm}
\end{figure}

We construct a multimodal network that learns independent representations for each modality and fuses layers at the end for multimodal classification.
We begin by building separate networks to learn low-dimensional representations for the spectral and image modalities. The spectral network, Fig.~\ref{fig:networks} (A), takes as input a 662-dimensional spectral sample and outputs material probability estimates from a softmax function. The model has two 64 node hidden layers followed by two 32 node layers, with batch normalization and a leaky ReLU activation applied after each layer. We apply a dropout of 0.25 after all but the last 32 node hidden layer.

Prior to training a model over texture images, we first feed images through a DenseNet-201 CNN pretrained on ImageNet~\cite{huang2017densenet}. We remove the 1000-class output layer such that the network outputs a feature vector of length 1920, resulting from global average pooling on the output of the preceding convolutional block. Models trained on ImageNet often learn strong representations for texture within an image~\cite{geirhos2018imagenettexture}. Given this, in Section~\ref{ssec:image_embeddings}, we compare material recognition results across various ImageNet-trained models used for computing texture image embeddings.

Fig.~\ref{fig:networks} (B) shows our image network, which takes as input the 1920-dimensional features from DenseNet-201. The network has three hidden layers of size 128, 64, and 32 nodes, with batch norm and leaky ReLU applied after each layer. We apply a dropout of 0.1 after the first two hidden layers. During evaluation (Section~\ref{sec:evaluation}), we train the spectral and image networks each for 50 epochs with a batch size of 128.

Given trained spectral and texture image models, we then define our multimodal network architecture.
We freeze the weights in both networks and remove the final 8 node output layer (depicted by the orange dotted lines in Fig.~\ref{fig:networks}). Both models output a 32-dimensional representation for their respective sensory modality. As depicted in Fig.~\ref{fig:networks} (C), these two outputs are concatenated and fed to a 32 node hidden layer
followed by a leaky ReLU activation. We use a softmax activation after the final 8 node output layer to compute probability estimates for each of the 8 material categories.
Since the spectral and texture image models are pretrained, we train only the weights for layers after the concatenation for 10 epochs.
We trained all models with the Adam optimizer, using $\beta_1=$~0.9, $\beta_2=$~0.999, and a learning rate of 0.0005.

Learning separate representations for each modality and combining into a shared representation for classification is commonly used for multimodal learning \cite{atrey2010multimodalsurvey, ngiam2011multimodal, eitel2015multimodalrgbd, liu2018combinemodalities}. 
From initial tests, we found that this late fusion approach performs better than directly learning a joint representation with early fusion. Overall, our results in Section~\ref{sec:evaluation} show that combining modalities improves generalization to recognize materials of unseen objects, with close-range texture imaging and near-field spectroscopy providing strong individual baselines.

\section{Evaluation}
\label{sec:evaluation}

\begin{figure}
\centering
\includegraphics[width=0.47\textwidth, trim={3cm 8cm 2cm 22cm}, clip]{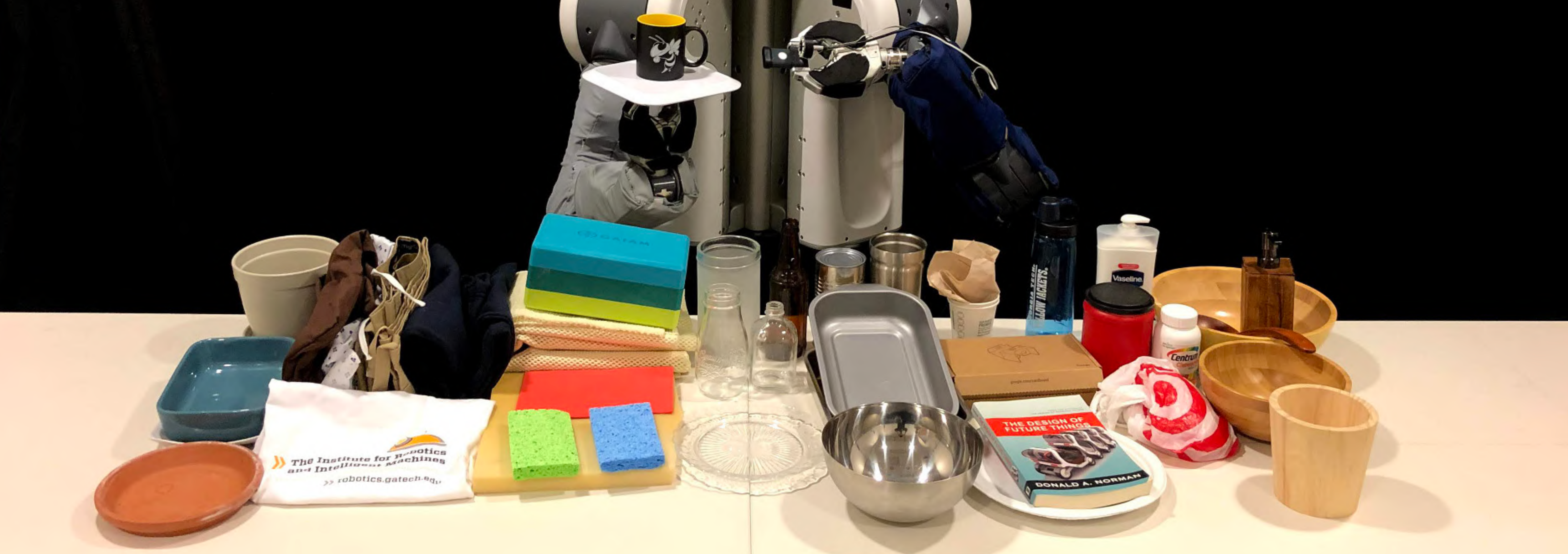}
\vspace{-0.2cm}
\caption{\label{fig:heldout_objects}The 40 heldout test set objects.}
\vspace{-0.7cm}
\end{figure}

Our dataset contains 14,400 spectral and image measurements from 144 distinct household objects. Prior to training and hyperparameter optimization for the models defined in Section~\ref{ssec:models}, we split these data into a training set of measurements from 104 objects, and a heldout test set of 40 objects (5 objects per each of the 8 material categories). This heldout data was not used for optimizing our models' hyperparameters. This heldout test set also includes the same test set objects used in~\cite{erickson2019spec}, shown in Fig.~\ref{fig:heldout_objects}, for a direct comparison to prior work that used only idealized spectral measurements\footnote{Idealized measurements are collected with flat material objects to block out environmental light and reduce noise in spectral measurements.}. To reduce the influence of random weight initialization when training models, we report all results averaged over 10 random seeds.

\subsection{Recognizing Materials of New Objects}

When deployed in real-world environments, robots are likely to encounter new objects which they have not yet been exposed to.
Similar to prior works in material classification, we begin by evaluating our multimodal sensing approach when recognizing the materials of new objects not found in the training data~\cite{erickson2019spec, erickson2017semi}. We first assess generalization across all 104 training set objects using leave-one-object-out cross-validation. To do so, we train a model on 103 objects (10,300 measurements) and evaluate material classification accuracy on the 100 samples from the one left-out object. We then repeat this process for each object and compute the average accuracy over the 104 splits.

As shown in Table~\ref{table:looo}, when using only spectral measurements with our spectral model (model A), we achieved an accuracy of 65.1\% averaged over 10 random seeds. When training on visual data, our image model (model B) achieved a material classification accuracy of 70.5\%. In comparison, our multimodal approach (model C) achieved an accuracy of 74.2\%, a $\sim$4\% improvement using low-dimensional representations of both image and spectral samples.

Prior research has investigated how a robot can use near-infrared spectroscopy to recognize object materials with leave-one-object-out cross-validation over five material categories: fabric, metal, paper, plastic, and wood~\cite{erickson2019spec}. For a direct comparison of results, we evaluate our performance given only these materials (excluding ceramic, foam, and glass objects). Table~\ref{table:looo} also depicts the performance of our models over these five materials (65 objects, 13 objects per material). Notably, our model trained on SCiO spectral measurements achieved 79.1\% accuracy, which is identical to the 79.1\% leave-one-object-out accuracy presented in prior work that used flat material objects~\cite{erickson2019spec}. 

As a final assessment of how spectroscopy and texture imaging enables generalizing material classification to new objects, we evaluate results over the heldout test set consisting of five objects from each material category. Prior work has evaluated how a model trained on idealized spectral measurements from flat objects can be used to recognize the materials of 25 household objects from five material categories (fabric, metal, paper, plastic, and wood)~\cite{erickson2019spec}. We include these same objects in our heldout test to enable a direct comparison with our multimodal sensing approach. \new{By training a multimodal model on both spectral and texture images from the training set (65 objects from five materials), our resulting model recognizes the materials of the 25 heldout test objects with 90.8\% accuracy, as shown in Table~\ref{table:looo_heldout}. When generalizing to new household objects, this is a $\sim$9\% improvement compared to the 81.6\% accuracy achieved in~\cite{erickson2019spec}, which trained a neural network model on only spectral measurements from flat material samples, rather than from household objects.} When compared to the leave-one-object-out results, we note that our multimodal approach performs significantly better on the paper and plastic heldout objects, 98.3\% and 77.6\% accuracy respectively, leading to higher overall performance on the heldout dataset.

\begin{table}
\centering
\caption{\label{table:looo}Leave-one-object-out accuracy with all 8 materials and the 5 materials from~\cite{erickson2019spec}.}
\begin{tabular}{cccc} \toprule
    & Spectral (A) & Image (B) & Multimodal (C) \\ \midrule\midrule
    5 Materials & \textbf{79.1} & 76.8 & \textbf{79.1} \\
    8 Materials & 65.1 & 70.5 & \textbf{74.2} \\
	\bottomrule
\end{tabular}
\end{table}

\begin{table}
\centering
\caption{\label{table:looo_heldout}Accuracy over the heldout test set, with all 8 materials and the 5 materials from~\cite{erickson2019spec}.}
\begin{tabular}{cccc} \toprule
    & Spectral (A) & Image (B) & Multimodal (C) \\ \midrule\midrule
    5 Materials & 85.9 & 80.1 & \textbf{90.8} \\
    8 Materials & 77.2 & 69.6 & \textbf{80.0} \\
\bottomrule
\end{tabular}
\vspace{-0.5cm}
\end{table}

\subsection{Spectral vs. Image Sensing}

In this section, we provide insight and case studies into what materials the two sensory modalities (spectral and image) perform best with and how a multimodal network architecture can leverage the strengths of each modality.

\begin{figure}
\centering
\raisebox{-.5\height}{\includegraphics[width=0.45\textwidth, trim={0cm 0cm 0cm 0cm}, clip]{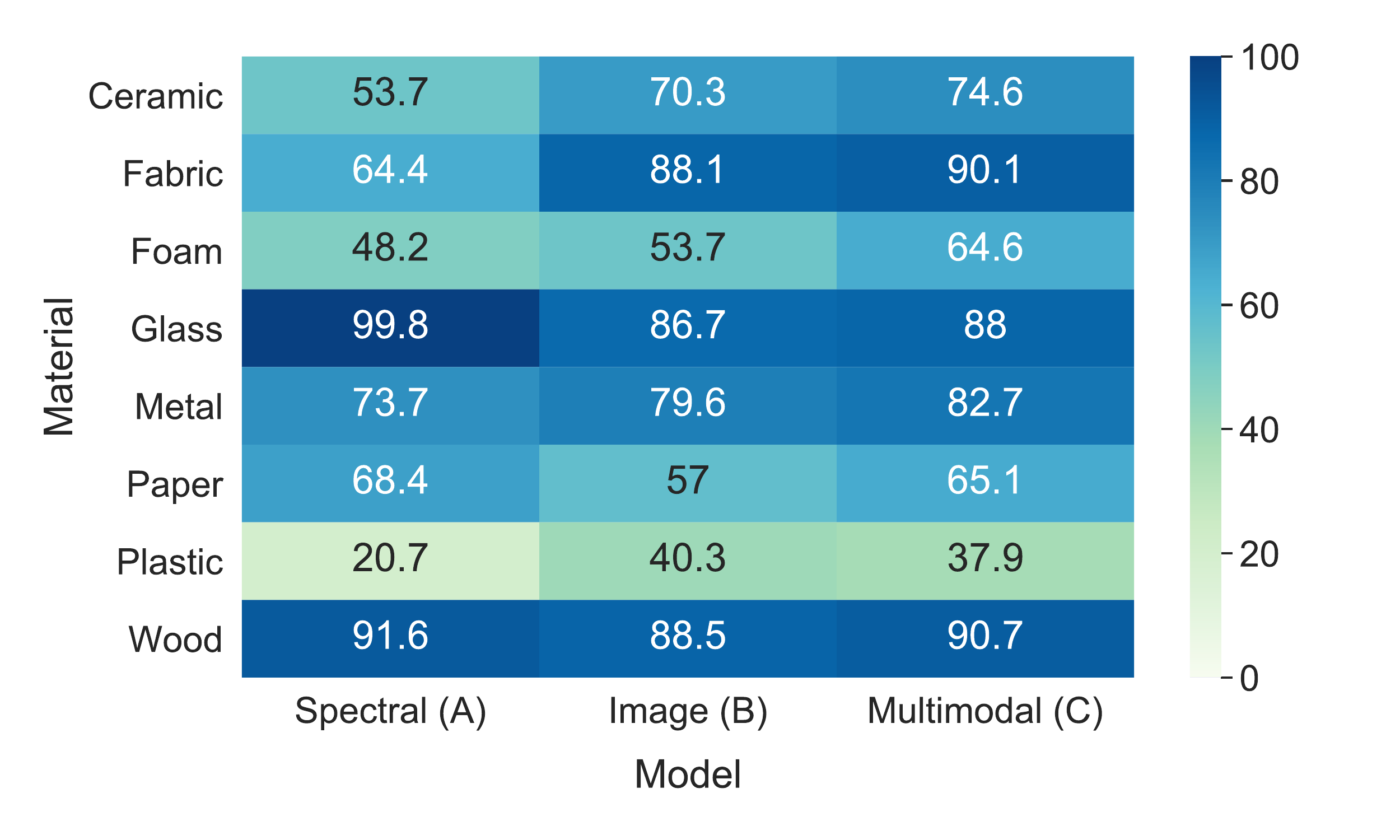}}
\vspace{-0.25cm}
\caption{\label{fig:looo_materials} Leave-one-object-out accuracy for each material.}
\vspace{-0.1cm}
\end{figure}

Fig.~\ref{fig:looo_materials} shows how our models trained on different modalities performed across material categories during leave-one-object-out cross-validation. We observe that it is easier to recognize fabrics with visual texture information, yet easier to recognize paper and glass with spectral data. Furthermore, some materials, such as plastic, remain difficult for both spectral and image data, in part due to large variation among plastic objects and difficulty distinguishing translucent plastics from glass.
In addition, we observe that in many cases, a multimodal model that leverages both spectral and visual data can more accurately recognize materials than when using either modality independently. One example of this occurring is with foam objects, where the spectral and image models achieved 48.2\% and 53.7\% accuracy, respectively, yet our multimodal model attained 64.6\% accuracy, $\sim$16\% higher than the spectral model and $\sim$11\% higher than the image modality.

\begin{figure}
\centering
\includegraphics[width=0.47\textwidth, trim={0cm 0.5cm 2cm 0cm}, clip]{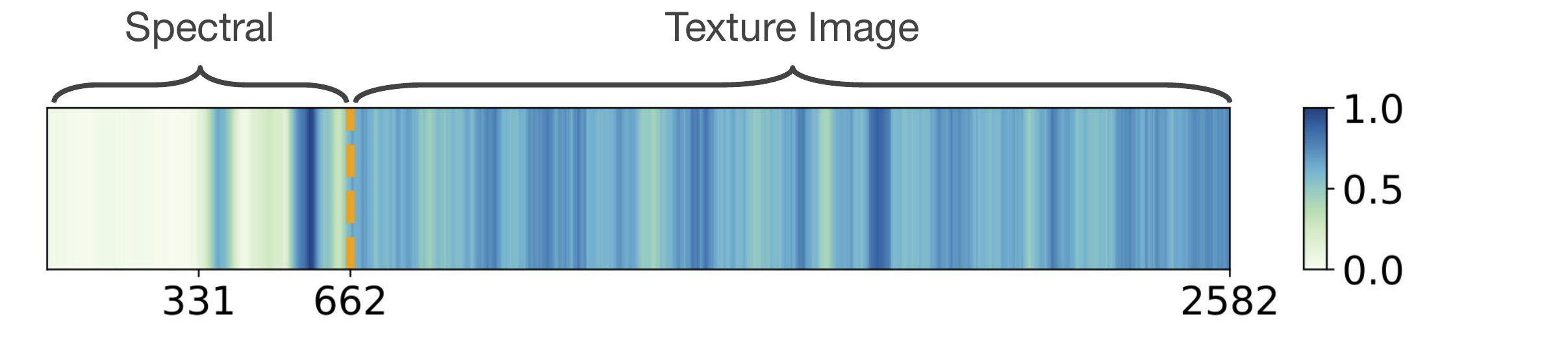}
\vspace{-0.1cm}
\caption{\label{fig:gradients}A saliency map from our multimodal model for a single spectral and image sample of the wood tray object. Gradients of the model output are backpropagated to the input vector to compute this saliency map. The range between 331 and 662 represents the difference quotient (derivative) of the spectral measurement.}
\vspace{-0.4cm}
\end{figure}

\begin{table}
\vspace{-0.2cm}
\centering
\caption{\label{table:looo_objects} Case studies of objects with leave-one-object-out accuracy. Images of each object are shown in Fig.~\ref{fig:imagewall}.}
\begin{tabular}{lcccc} \toprule
    Object & Spectral (A) & Image (B) & Multimodal (C) \\ \midrule\midrule
    Fabric gray shirt & 2.4 & 99.7 & 99.7 \\
    Foam plate & 100.0 & 43.3 & 99.4 \\
    Wood tray & 0.2 & 63.6 & 78.4 \\
    Plastic coffee-mate & 0.6 & 81.5 & 50.5 \\
    Paper tissue box & 100.0 & 15.8 & 48.2 \\
	\bottomrule
\end{tabular}
\end{table}

\begin{figure*}
\centering
\includegraphics[width=0.4\textwidth, trim={2.95cm 0cm 0cm 0cm}, clip]{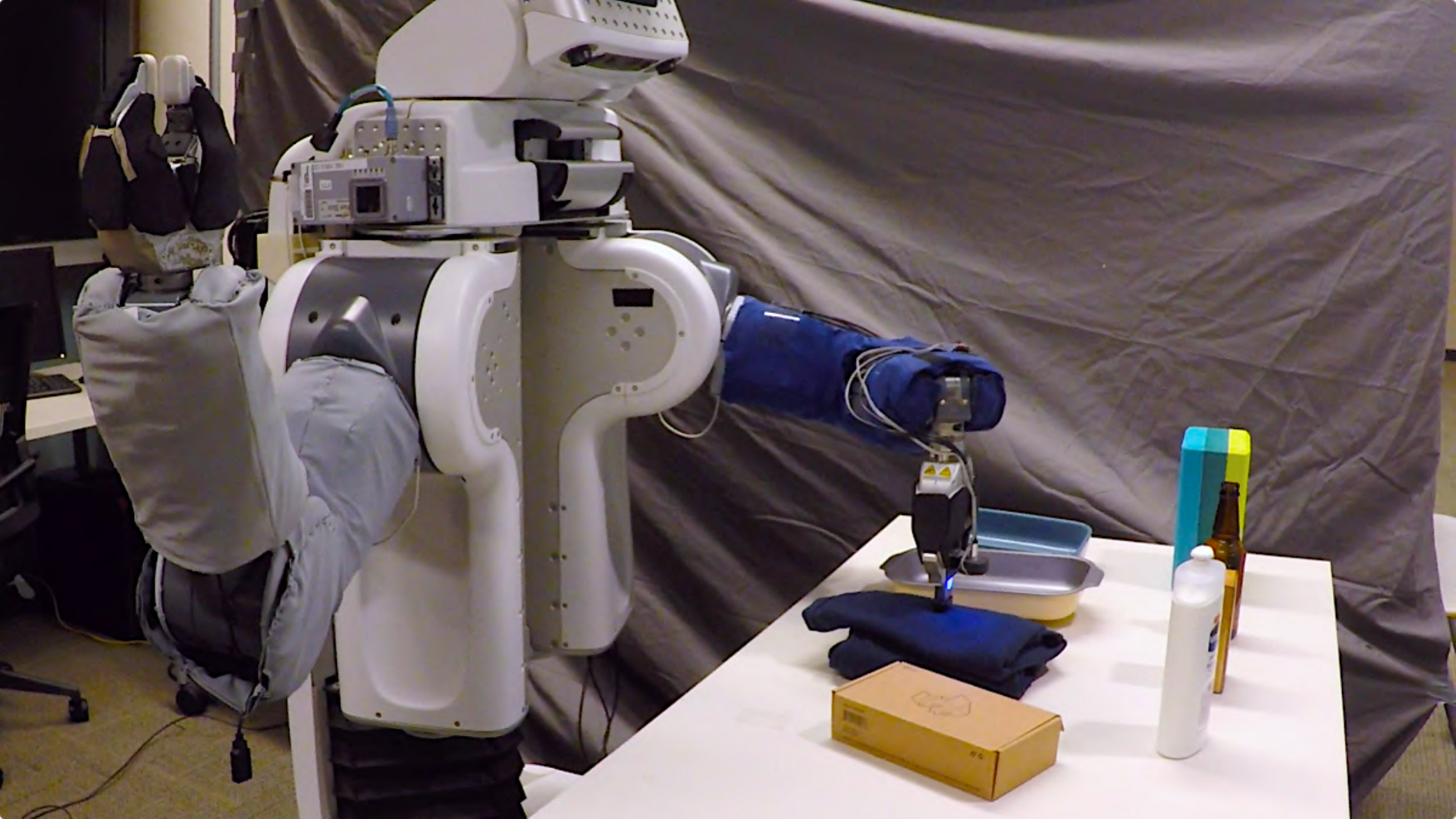} 
\hspace{1cm}
\includegraphics[width=0.4\textwidth, trim={0cm 0cm 0cm 0cm}, clip]{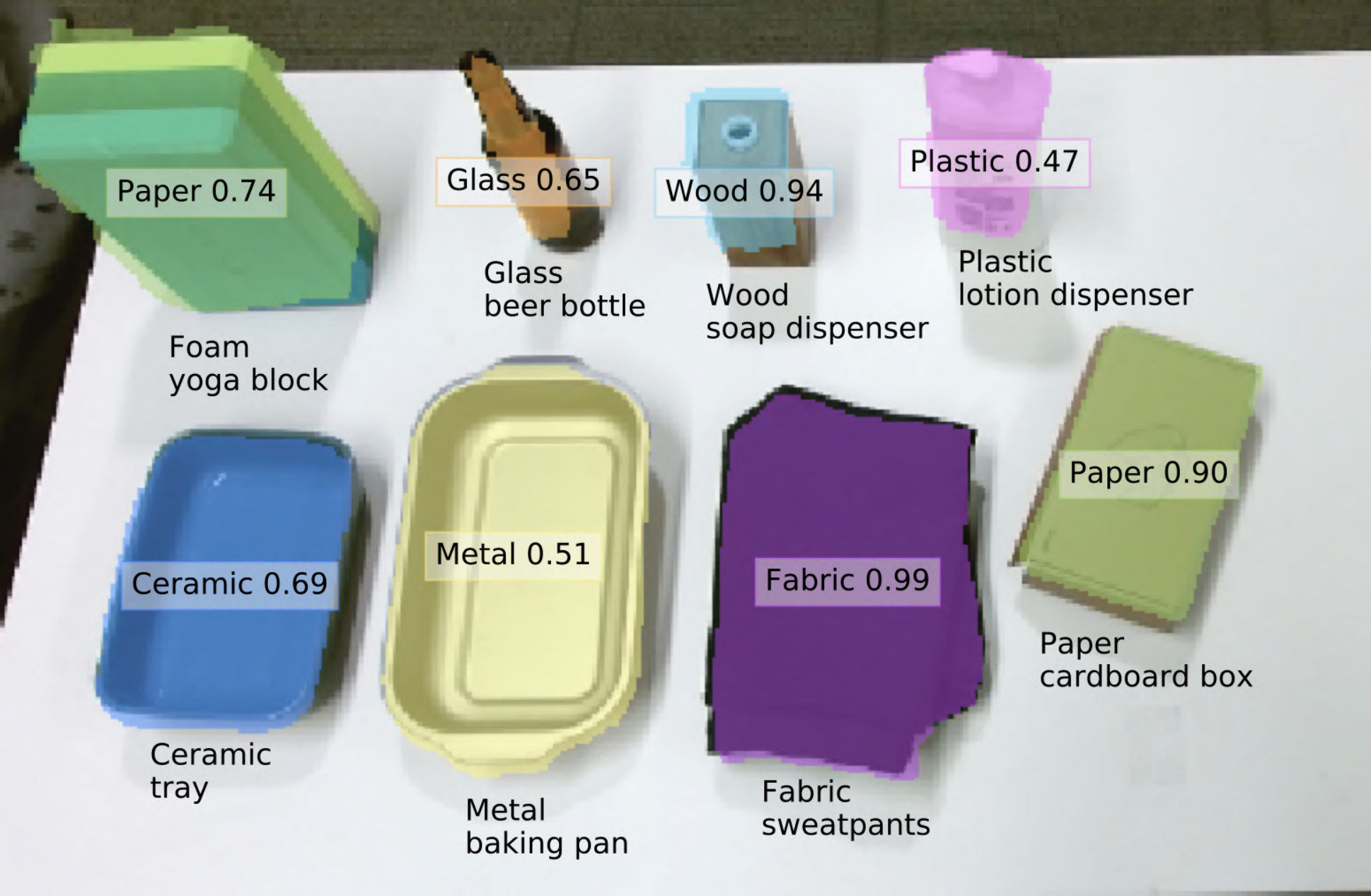}
\vspace{-0.2cm}
\caption{\label{fig:evaluation}Demonstration of object material recognition. (Left) Table scene of objects. (Right) Pixel-level material classification with accompanying prediction probabilities using our multimodal approach.}
\vspace{-0.5cm}
\end{figure*}

Beyond averages over entire material categories, we also investigate examples of specific objects and how the different modalities compare, as shown in Table~\ref{table:looo_objects}. A gray cotton fabric shirt was challenging for our spectral model to classify, achieving only 2.4\% accuracy. In comparison, the image modality recognized this object as fabric with 99.7\% accuracy. Our multimodal model also achieved 99.7\% accuracy by learning to leverage the visual information to make its decision.
Conversely, our image model struggled to accurately classify a foam plate with only 43.3\% accuracy. By incorporating spectral data, our multimodal model correctly recognizes the foam plate with 99.4\% accuracy.

As indicated in the previous examples, a multimodal model frequently matches or outperforms models trained on independent modalities.
Another example of this is a wood tray for which the spectral and texture image models reach 0.2\% and 63.6\% accuracy, respectively. Yet our multimodal model recognizes this object as wood with 78.4\% accuracy, a $\sim$15\% improvement over the image model.
Fig.~\ref{fig:gradients} shows a saliency map from our multimodal model (visualization of what input features would affect the material estimate most if changed)~\cite{simonyan2013deep} for a single measurement from the wood tray.
As depicted, our multimodal model uses the entire image modality to make its classification, but also uses a small portion of the spectral data to further improve its estimate.

A few limitations still remain with using a multimodal network architecture.
Namely, there are instances where using a model trained on either spectral or image modalities independently performs better than a model trained on both modalities.
This phenomenon occurs with both the plastic coffee-mate container and paper tissue box objects during leave-one-object-out cross-validation. For the plastic coffee-mate, texture imaging alone achieved 81.5\% accuracy over 10 random seeds. Yet, our multimodal model only recognized this object as plastic 50.5\% of the time.
Similarly, our spectral model recognized the paper tissue box with 100.0\% accuracy, yet when combined with image data, our multimodal model only classified this as paper 48.2\% of the time.
Upon inspection, we observed that the image model incorrectly classified 83.3\% of paper tissue box images as plastic. We see these tissue box images often contain text similar to labels on commercial plastic objects in the dataset. These inaccuracies may be due in part to our dataset size, which may not fully capture the variation of materials across real-world objects. Increasing the number and variety of objects in the training set may further improve performance when generalizing material classification to new objects.

\subsection{Texture Image Features}
\label{ssec:image_embeddings}

\begin{table}
\centering
\caption{\label{table:image_preprocessing}8 material leave-one-object-out accuracy (model B), resizing or center cropping image input for DenseNet-201 features.}
\begin{tabular}{lc} \toprule
    Image Preprocessing & Accuracy \\ \midrule\midrule
    $(320 \times 240)$ resize & \textbf{70.5} \\
    $(320 \times 240)$ crop & 61.6 \\
    $(640 \times 480)$ resize & 69.0 \\
    $(640 \times 480)$ crop & 65.7 \\
    $(1280 \times 960)$ resize & 66.2 \\
	\bottomrule
\end{tabular}
\vspace{-0.5cm}
\end{table}

In Section~\ref{sec:dataset}, we presented a neural network architecture for material classification with visual texture data, dependent on preprocessing images with a DenseNet-201 model trained on ImageNet. Prior research has indicated that models trained on ImageNet have a high prior for recognizing texture features within an image~\cite{geirhos2018imagenettexture}. In this section, we evaluate different texture image preprocessing techniques and compare ImageNet-trained models for texture representation.

The original resolution of the captured close-range texture images is $1600 \times 1200$. CNN models are usually trained on ImageNet using center crops of $224 \times 224$. We test different resolutions for our texture images by center crop or resize, using DenseNet-201 as a feature extractor. Results are shown in Table~\ref{table:image_preprocessing} for different raw image preprocessing techniques prior to computing the DenseNet-201 features, evaluated on 8 material leave-one-object-out material classification with the texture image model (model B). 
We find that resizing performs better than center cropping, suggesting that textural features are better captured with more visual surface area and context of the object, rather than a small but dense visual sample.
Additionally, we observe that resizing to $320 \times 240$, which is near the image resolution that the CNN was trained at, performs better than resizing to higher resolutions. 

\begin{table}
\centering
\caption{\label{table:imagenet_models}Leave-one-object-out accuracy (model B), comparing features from ImageNet models. $(320 \times 240)$ resized images.}
\begin{tabular}{lc} \toprule
    Network & Accuracy \\ \midrule\midrule
    VGG19~\cite{simonyan2014vgg} & 63.7 \\
    ResNet-50~\cite{he2016resnet} & 66.2 \\
    ResNet-101~\cite{he2016resnet} & 67.4 \\
    ResNet-152~\cite{he2016resnet} & 66.0 \\
    DenseNet-201~\cite{huang2017densenet} & \textbf{70.5} \\
    ResNeXt-101~\cite{xie2017resnext} & 68.7 \\
    \footnotemark[3]EfficientNet-B5~\cite{tan2019efficientnet} & 69.4 \\ 
	\bottomrule
\end{tabular}
\vspace{-0.5cm}
\end{table}

We generate low-dimensional visual features from common ImageNet-trained CNNs using texture images that were resized\footnote{Texture images were resized to $608 \times 456$ for EfficientNet-B5, near its native ImageNet input resolution.} to a resolution of $320 \times 240$.
Table~\ref{table:imagenet_models} compares several ImageNet models for computing texture representations, which are then used to train the texture image model (model B) during leave-one-object-out cross-validation on all 8 materials.
We observe that performance on ImageNet is loosely correlated with material classification accuracy with our image model. Further advances on CNN models benchmarked by ImageNet may continue to improve texture representation. Due to the architecture of our multimodal network, which learns separate and combined representations, advances in texture representation should lead to improvements on material classification with texture images.

\subsection{Table Scene Recognition}

We further evaluate our multimodal sensing approach by classifying materials of a scene of objects placed on a table, similar to what may be observed in a kitchen or home environment. We place one object from each material category from the heldout dataset on a table in front of the PR2. Using a 3D point cloud from its head-mounted Kinect, the PR2 segments objects from the table and defines pixel-level clusters in the 2D visual image for each distinct object found in the point cloud.
The robot then classifies the material of each cluster using spectral and close-range texture image measurements from each object.
To capture a measurement, the PR2 moves its left end effector to a position just in front of each object, matching a surface normal for the object computed from the point cloud.
Fig.~\ref{fig:evaluation} shows a table setup with the PR2 and the pixel-level classification of each object using predictions from our multimodal material classification model. Our model correctly recognized the materials for seven of the eight heldout objects, missing only the foam yoga block. This demonstration can be seen in greater detail in the supplementary video.

\section{Conclusion}

This paper introduces a multimodal sensing technique that combines near-infrared spectroscopy and close-range high resolution texture imaging for enabling robots to accurately classify the materials of household objects.
We present and evaluate a new dataset of spectral measurements and high-resolution texture images for 144 household objects from 8 material categories. Compared to prior work in material classification with spectroscopy, our multimodal approach achieved 9\% higher accuracy when generalizing to new, unseen household objects.
In addition, we demonstrate how this sensing technique enables a robot to recognize materials across a scene of objects on a table, without physical contact with the objects.
Through this work, we have shown that near-infrared spectroscopy and texturing imaging offers a reliable and accurate multimodal sensing approach for robots to estimate the materials of objects.

\bibliographystyle{IEEEtran}
\bibliography{bibliography}

\end{document}